\documentclass{amia}
\usepackage{graphicx}
\usepackage[labelfont=bf]{caption}
\usepackage[superscript,nomove]{cite}
\usepackage[dvipsnames]{xcolor}
\usepackage[sort&compress,comma,numbers,super]{natbib}
\usepackage{multirow}
\usepackage{url}
\usepackage{soul}
\usepackage{hyperref}
\usepackage{listings}
\usepackage{amsmath}
\usepackage{needspace}

\usepackage{booktabs}
\usepackage{multirow}
\usepackage{makecell}
\usepackage{array}
\usepackage{pifont}
\usepackage{bm}
\usepackage{relsize}
\usepackage{amsmath}
\usepackage{subcaption}
\usepackage{sidecap}

\newcommand\EatDot[1]{}

\definecolor{myblue}{HTML}{4E79A7}
\definecolor{mygreen}{HTML}{59A14F}
\definecolor{myred}{HTML}{E15759}

\definecolor{myteal}{HTML}{75b6b2}

\definecolor{highlight-yellow}{HTML}{ffff00}
\definecolor{highlight-green}{HTML}{00ff00}
\definecolor{highlight-teal}{HTML}{00ffff}

\newcommand{\hlc}[2][yellow]{{%
    \colorlet{foo}{#1}%
    \sethlcolor{foo}\hl{#2}}%
}

\newcolumntype{P}[1]{>{\centering\arraybackslash}p{#1}}
\newcolumntype{M}[1]{>{\centering\arraybackslash}m{#1}}

\newcolumntype{-}{>{\global\let\currentrowstyle\relax}}
\newcolumntype{^}{>{\currentrowstyle}}
\newcommand{\rowstyle}[1]{\gdef\currentrowstyle{#1}
  #1\ignorespaces
}

\newcommand*\dataset{\textsc{ChartPNG}}
\mathchardef\mhyphen="2D % Define a "math hyphen"

\begin{document}

\title{Toward Relieving Clinician Burden by Automatically Generating Progress Notes using Interim Hospital Data}

\author{Sarvesh Soni, PhD, Dina Demner-Fushman, MD, PhD}

\institutes{
    Lister Hill National Center for Biomedical Communications\\
    National Library of Medicine, National Institutes of Health\\
    Bethesda, MD, USA
}

\maketitle

\noindent{\bf Abstract}

\textit{
Regular documentation of progress notes is one of the main contributors to clinician burden.
The abundance of structured chart information in medical records further exacerbates the burden, however, it also presents an opportunity to automate the generation of progress notes.
In this paper, we propose a task to automate progress note generation using structured or tabular information present in electronic health records.
To this end, we present a novel framework and a large dataset, \dataset, for the task which contains $7089$ annotation instances (each having a pair of progress notes and interim structured chart data) across $1616$ patients.
We establish baselines on the dataset using large language models from general and biomedical domains.
We perform both automated (where the best performing Biomistral model achieved a BERTScore F1 of $80.53$ and MEDCON score of $19.61$) and manual (where we found that the model was able to leverage relevant structured data with $76.9\%$ accuracy) analyses to identify the challenges with the proposed task and opportunities for future research.
}

\section{Introduction}

Progress notes, also known as SOAP notes, are written by physicians in electronic health records (EHRs) and contain two broad categories of content: (1) Subjective and Objective status of a patient and (2) Assessment and Plan (A\&P) \cite{pearce2016EssentialSOAPNote}.
These notes are written at regular intervals (e.g., every $24$ hours) to document the journey of patient care for monitoring, sharing care responsibilities, and record-keeping.
Documenting progress notes is the main contributor to physicians spending almost half of their time in front of computers, highlighting the need to automate this task \cite{tai-seale2017ElectronicHealthRecord}.

The patient information is continuously collected (e.g., laboratory values, mental status, equipment settings) during the provision of care, especially during a hospital stay, and stored in the structured or tabular format in the patient charts.
On one hand, the abundant patient chart information present in EHRs enables physicians to better assess the progress of their patients (and thus write better EHR notes), while on the other hand, this presents associated challenges related to information overload and increased documentation burden.
To this end, it is natural to make use of the available structured EHR data for automating progress note generation (PNG).
However, most existing work on PNG focused on using doctor-patient conversations as input \cite{ramprasad2023GeneratingMoreFaithful,benabacha2023EmpiricalStudyClinical,krishna2021GeneratingSOAPNotes}, likely due to the ready availability of such data for the task.
However, to our knowledge, no previous study harnessed the structured chart data available in the EHRs for PNG.

The Subjective and Objective sections of a progress note are based on the information collected about the patient.
Information in the Subjective section is the patient's (or their family's) interpretation of their condition, thus it needs to be derived from the physician-patient conversation itself.
Information in the Objective section is oftentimes pulled directly from the patient charts into the note as it is and relatively trivial to automate as its assembly does not require substantial human effort.
Differently, the content in the A\&P sections is composed by physicians after carefully examining the relevant patient information (including previous notes and structured data).
Thus, in this study, we focus on automatically generating the A\&P sections based on the content already available in the EHR (i.e., past progress notes and structured chart data).
The main contributions of this study are as follows.

\begin{figure}[t]
    \begin{center}
        \includegraphics[width=\columnwidth]{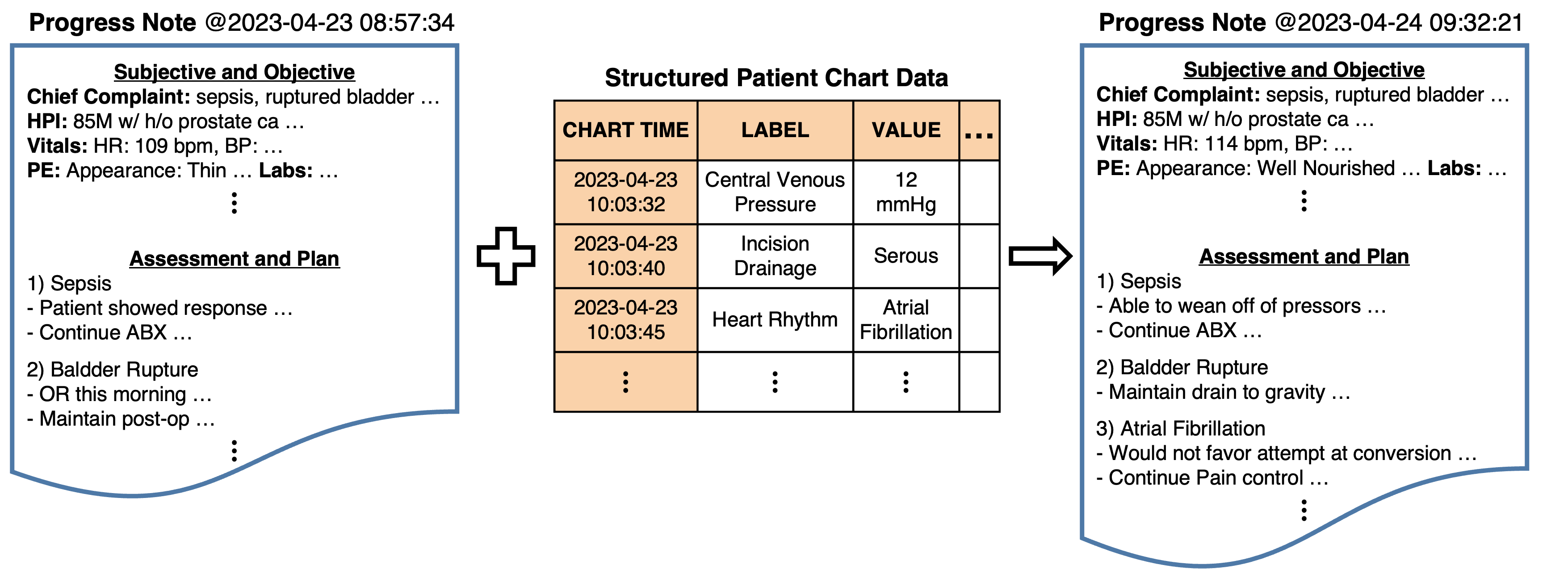} 
        \caption{Overview of the proposed task to automatically generate the next progress note using the previous note and all interim structured chart data.}
        \vspace{-9pt}
        \label{fig:system}
    \end{center}
\end{figure}

\begin{itemize}
    \item Introduce a novel method for automatically generating progress notes using available structured patient chart data (Figure \ref{fig:system}).
    \item Propose a novel dataset, \dataset, for generating progress notes using structured EHR data\footnote{\href{https://github.com/soni-sarvesh/progress-note-generation}{github.com/soni-sarvesh/progress-note-generation}}.
    \item Identify the challenges involved in PNG by conducting an analysis of errors made by the baseline models.
\end{itemize}

\section{Related Work}

While the task of PNG is relatively unexplored in light of structured data, both automated PNG and structured data-to-text are well-researched.
In the subsequent sections, we discuss the advancements made in both areas and further highlight the novelty of our approach in the context of current research.

\vspace{-6pt}

\subsection{Progress Note Generation}

Automated text generation is a long-standing problem in medicine \cite{cawsey1997NaturalLanguageGeneration} with several efforts specifically focused on progress notes.
However, most past studies required clinicians to enter specific note-related information (including text snippets) into the system before a note was assembled using templates \cite{fischer1987UserReactionPROMIS,ma1991InteractiveSystemGenerating,lowe1992ComprehensiveComputerizedNeonatal,campbell1993ComputerbasedToolGeneration,huske-kraus2003Suregen2ShellSystem,harris2008BuildingLargescaleCommercial}.
Thus, these approaches still burden clinicians with the task of gathering several important pieces of the notes.

Several studies have focused on PNG using medical conversations as input.
Krishna et al. \cite{krishna2021GeneratingSOAPNotes} proposed an algorithm for PNG for conversations from doctor-patient visits using deep summarization models and conducted evaluations on their proprietary medical dataset.
Ramprasad et al. \cite{ramprasad2023GeneratingMoreFaithful} used a similar proprietary conversations dataset for improving PNG using section-specific adapters.
Yim et al. \cite{yim2021AutomatingMedicalScribing} evaluated several summarization models for generating note snippets from medical visit conversations, which additionally included statements and commands intended for the medical scribes to expand those into the scribed notes.
Michalopoulos et al. \cite{michalopoulos2022MedicalSumGuidedClinical} generated progress notes from Family Medicine doctor-patient conversations using transformer-based models integrated with medical knowledge.
Enarvi et al. \cite{enarvi2020GeneratingMedicalReports} compared the transformer and recurrent neural network-based models for the task of PNG for visit dialogue from ambulatory orthopedic surgery encounters.
Note that all these studies are conducted for outpatient setting data, where it is conceivable that A\&P can be collected from the medical conversations.
On the contrary, our work focuses on the inpatient setting where A\&P are more complex (\textit{``Atrial Fibrillation: Lopressor IV, Dilt as needed for HR control''}) and are likely difficult to capture directly from the medical conversations in this setting where sometimes a medical dialogue may not be possible altogether (e.g., the patient is unable to communicate).
On the flip side, a tremendous amount of structured chart data is continuously collected during in-patient encounters, making it available for automating PNG.

Interestingly, much work on automated note generation focused on generating the Subjective section (or its subset) of the progress notes from doctor-patient conversations.
Zhang et al. \cite{zhang2021LeveragingPretrainedModels} centered on only the History of Present Illness (HPI) section of the progress notes as they observed that the other note sections were less frequently extracted from the input doctor-patient conversations.
Similarly, Joshi et al. \cite{joshi2020DrSummarizeGlobal} and Chintagunta et al. \cite{chintagunta2021MedicallyAwareGPT3} extracted past medical history information (e.g., symptoms, patient concerns, psychological and social history) from medical conversations.
Liu et al. \cite{liu2019TopicAwarePointerGeneratorNetworks} extracted symptom information (corresponding to \textit{Review of Systems} subsection) from the nurse-patient telemonitoring dialogue.
Moramarco et al. \cite{moramarco2022HumanEvaluationCorrelation,moramarco2021PreliminaryStudyEvaluating} systematically evaluated the generation of Subjective sections from consultation conversations.
This further bolsters that it is challenging to extract all the sections from medical conversations alone.

Further, shared tasks for dialogue to note generation, namely, MEDIQA-Chat \cite{benabacha2023OverviewMEDIQAChat2023} and MEDIQA-Sum \cite{yim2023OverviewMEDIQASumTask}, provided publicly available datasets on the task which facilitated several research efforts in this direction.
Here, one dataset, MTS-Dialog \cite{benabacha2023EmpiricalStudyClinical}, was synthetically created against a given set of clinical notes (including SOAP-based notes) while another, ACI-BENCH \cite{yim2023AcibenchNovelAmbient}, was constructed through role-playing.
In a similar spirit, our dataset is publicly available, laying out the foundations for future research.

\subsection{Structured Data for Text Generation}

Text generation is a widely studied field \cite{gatt2018SurveyStateArt,fatima2022SystematicLiteratureReview,lin2023SurveyNeuralDatatoText}.
However, fewer advances have been made in the healthcare domain \cite{pauws2019MakingEffectiveUse}, with most approaches making use of structured data entries and canned text \cite{huske-kraus2003TextGenerationClinical}.
Two systems are developed to generate nursing shift summaries from the Neonatal Intensive Care Unit (NICU) data, as part of the BabyTalk project, for different durations of data such as 45 minutes (BT-45, \cite{portet2009AutomaticGenerationTextual}) and 12 hours (BT-Nurse, \cite{hunter2011BTNurseComputerGeneration}).
However, the nursing summaries and progress notes differ in their syntax and semantics and thus need to be explored independently.
Much work is present on automatically generating radiology reports from radiological examination images \cite{monshi2020DeepLearningGenerating,messina2022SurveyDeepLearning,ramirez-alonso2022MedicalReportGeneration,liao2023DeepLearningApproaches,pang2023SurveyAutomaticGeneration}.
Nonetheless, it is imperative to recognize the inherent differences between the data modalities (tabular versus imaging).

\section{\dataset\ Dataset}

We curated the proposed corpus using data from MIMIC-III \cite{johnson2016MIMICIIIFreelyAccessible}, a publicly available critical care database.
Each annotation instance in \dataset\ consists of a pair of progress notes along with all interim structured chart data (between the documentation times of the \textit{prior} and the \textit{next} notes in a pair).
The first (\textit{prior}) note in the pair serves as an input to large language model (LLM) along with structured data while the second (\textit{next}) note in the pair serves as gold standard (ground truth).
Progress notes are written by different types of clinicians such as Attending Physicians, Medical Residents, and Nurses and, thus, they serve different purposes and differ in content as well as writing style.
We focus on the progress notes written by Attending Physicians as they are responsible for a patient's care in the hospitals.
The pairs of notes were selected if they (1) belong to the same admission and, between their documentation times, there is (2) no other documented progress note and (3) non-empty structured chart data.

\begin{table}[t]
\caption{Descriptive statistics of the curated dataset, \dataset.} \label{tab:data_stats}
	\centering
	\def\arraystretch{1.15}
	\begin{tabular}{l c c c}
            \hline
            \textbf{Item} & \multicolumn{3}{c}{\textbf{Count}} \\
            \hline
            Patients & \multicolumn{3}{c}{$1616$} \\
            Annotation instances (note pairs) & \multicolumn{3}{c}{$7089$} \\
            $\hookrightarrow$ Mean instances / patient & \multicolumn{3}{c}{$4.4$} \\
            $\hookrightarrow$ Median instances / patient & \multicolumn{3}{c}{$2$} \\
            \hline
            & \textbf{Mean} & \textbf{Median} & \textbf{SD} \\
            \hline
            Structured chart data (rows) & $1474.9$ & $1204$ & $1569.4$ \\
            Prior A\&P length (words) & $203.5$ & $177$ & $124.2$ \\
            Next A\&P length (words) & $198.2$ & $171$ & $123.6$ \\
            $\hookrightarrow$ when text added (total: $3551$) & $235.2$ & $207$ & $132.2$ \\
            $\hookrightarrow$ when text reduced (total: $3551$) & $161.3$ & $140$ & $101.8$ \\ \hline
        \end{tabular}
\end{table}

Table \ref{tab:data_stats} presents the statistics of \dataset.
Note that there is a huge amount of structured data between the two progress notes in each annotation instance with an average of $1474.9$ rows.
The prior progress notes are as long as $203.5$ words on average which gets more challenging due to the frequent use of telegraphic text in EHR notes by Physicians, which packs more information in lesser words.
Understandably, the distribution of prior and next progress notes are similar.
However, the distributions of the next notes stratified by the addition or removal of text from the prior note indicates that the next notes differ in length from the prior notes by roughly $35$ words (through addition and/or removal of information).

\section{Methods}

\begin{figure}[t]
    \begin{center}
        \includegraphics[width=\columnwidth]{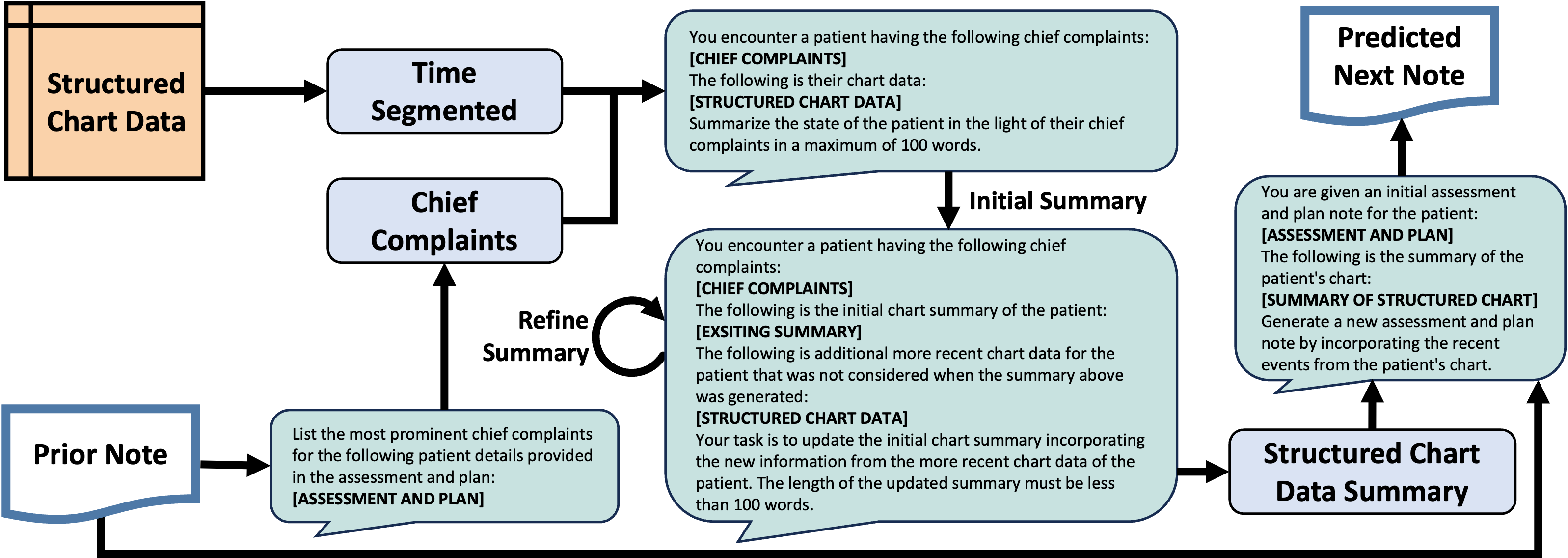} 
        \caption{Overview of the proposed framework for PNG. Prompts used for the generative large language models are shown in \textbf{\color{myteal} teal} chat boxes.}
        \label{fig:methods}
    \end{center}
\end{figure}

Our proposed framework for the task of PNG is presented in Figure \ref{fig:methods}.
We used open-source LLMs from general as well as biomedical domains as base models for our proposed task.
The use of proprietary or closed-source LLMs raises several privacy and security concerns in healthcare, especially in the context of PNG \cite{nguyen2023ApplicationChatGPTHealthcare}.
Since we are dealing with sensitive data (from MIMIC-III), in a similar vein, we investigate locally-executable open-source models for the task.
Specifically, we reported results from Biomistral 7B \cite{labrak2024BioMistralCollectionOpenSource}, Mixtral 8x7B \cite{jiang2024MixtralExperts}, and LLaMa 2 70B \cite{touvron2023LlamaOpenFoundation} models.
Biomistral is based on the Mistral model that is further pre-trained on the PubMed Central Open Access Subset.
We chose Biomistral as it surpassed the performance of existing open-source medical models on multiple tasks \cite{labrak2024BioMistralCollectionOpenSource}.
The Mixtral model is an improved variation of the Mistral model and has outperformed several state-of-the-art LLMs on different tasks \cite{jiang2024MixtralExperts}.
The LLaMa 2 model is the second generation variant of the LLaMa models and has been shown to outperform the proprietary models on some tasks \cite{touvron2023LlamaOpenFoundation}.

The lengthy tabular chart data (see Table \ref{tab:data_stats}) is difficult to accommodate by the small context size of most open-source medical models including Biomistral which can handle up to $2k$ tokens (about $1250$ words including model input and output texts).
LLaMa 2 and Mixtral have longer context lengths, however, since the main focus of our evaluation is to establish fair baselines rather than achieving the state-of-the-art performance we use the context length of $2k$ tokens for all models.
Thus, we summarize the structured chart data for the purposes of PNG.
In order to further save prompt real estate, we condense the tabular data by mentioning each timestamp once and listing associated data thereafter.
To direct the summarization process, we extract patient's current chief complaints from the A\&P section of the prior note (as opposed to using the broad chief complaints from admission).
Due to the context size limitations, the summarization step is also carried out in two iterative steps where an initial summary is obtained using the first chunk of the tabular chart data and the subsequent chunks are used to refine this existing summary.
Finally, the prior A\&P note and the synthesized summary of the structured EHR data is used to generate the next A\&P note.
The detailed information flow along with the model input prompts are shown are Figure \ref{fig:methods}.

We report automated metrics for evaluating text generation, namely, ROUGE \cite{lin2004ROUGEPackageAutomatic}, BERTScore \cite{zhang2019BERTScoreEvaluatingText} (using the model RoBERTa$_{\textsc{large}}$ \cite{liu2019RoBERTaRobustlyOptimized}), and MEDCON \cite{yim2023AcibenchNovelAmbient}, by comparing the predicted A\&P notes to ground truth next A\&P notes.
ROUGE measures the text overlaps while BERTScore takes into account the semantic similarity between gold and prediction.
Differently, MEDCON extracts Unified Medical Language System (UMLS) concepts from both gold and predicted notes and calculates an F1-score between the two sets of concepts to measure similarity.
As a baseline in the evaluation, the prior note in the pair that serves as input to LLM along with structured data is used with no changes.
We additionally performed a manual evaluation on a randomly selected sample.
Here, we manually classified the predicted progress notes into different categories and analyzed the results with examples.

\section{Results}

\begin{table}[t]
\caption{Evaluation results with multiple baselines. Larger models, Mixtral 8x7B and LLaMa 2 70B, are evaluated on $30$ annotation instances due to hardware constraints. \textit{Prior} -- output the prior note verbatim as prediction.} \label{tab:baseline_results}
	\centering
	\def\arraystretch{1.15}
	\begin{tabular}{c@{\hspace{10pt}}c@{\hspace{10pt}}c@{\hspace{10pt}}c@{\hspace{10pt}}c@{\hspace{10pt}}cc@{\hspace{10pt}}c@{\hspace{10pt}}c@{\hspace{10pt}}cc@{\hspace{10pt}}c}
            \hline
            \multirow{2}{*}{\textbf{Baseline}} & & \multicolumn{4}{c}{\textbf{ROUGE}} & & \multicolumn{3}{c}{\textbf{BERTScore}} & & \textbf{MEDCON} \\
            \cline{3-6} \cline{8-10} \cline{12-12}
             & & \textbf{1} & \textbf{2} & \textbf{L} & \textbf{Lsum} & & \textbf{Precision} & \textbf{Recall} & \textbf{F1} & & \textbf{F1} \\
            \hline
            Prior & & $50.74$ & $35.40$ & $42.06$ & $50.21$ & & $89.12$ & $89.13$ & $89.11$ & & $55.52$ \\
            Biomistral 7B & & $18.90$ & $5.20$ & $11.47$ & $17.95$ & & $81.44$ & $79.71$ & $80.53$ & & $19.61$\\
            \hline
            \multicolumn{12}{c}{\textbf{On the subsample used for manual evaluations}}\\
            \hline
            Prior & & $51.77$ & $35.04$ & $42.13$ & $50.73$ & & $89.62$ & $89.94$ & $89.77$ & & $55.46$ \\
            Biomistral 7B & & $20.85$ & $7.36$ & $13.81$ & $19.88$ & & $82.08$ & $80.87$ & $81.42$ & & $21.99$ \\
            Mixtral 8x7B & & $20.29$ & $3.84$ & $10.99$ & $19.19$ & & $80.96$ & $78.44$ & $79.66$ & & $20.04$ \\
            LLaMa 2 70B & & $16.75$ & $2.64$ & $8.97$ & $16.03$ & & $80.21$ & $77.68$ & $78.91$ & & $16.75$ \\
            \hline
        \end{tabular}
\end{table}

Automated performance metrics from our baselines are reported in Table \ref{tab:baseline_results}.
The simple baseline of returning the prior note as prediction resulted in the highest scores across the automated metrics.
Interestingly, this is an artifact of the high textual and semantic similarity between gold prior and next notes as, oftentimes, information text is copied between the progress notes.
The low ROUGE scores from the LLM predictions underscores the limited overlap of text between the predicted and gold next A\&P notes, as also seen in our manual analysis (Table \ref{tab:manual_eval_with_eg}).
Among the evaluated LLMs, Biomistral achieved the best BERTScore F1 of $81.42$ and MEDCON score of $21.99$, which could be attributed to the medical knowledge that the model acquired through further training on the biomedical resources.

\begin{table}[t]
\caption{Common prediction characteristics from a manual evaluation of the models predictions on $30$ annotation instances. Values in $\%~(\#)$ format.}  \label{tab:manual_eval_with_eg}
	\centering
	\setul{0.25ex}{}
	\def\arraystretch{1.25}
	\setlength\tabcolsep{3pt}
	\resizebox{\linewidth}{!}{
	\begin{tabular}{
	    >{\raggedright}-m{0.11\linewidth}
	    ^m{0.3\linewidth}
	   ^l
          ^c
	   ^c
          ^c
	}
		\hline 
		 \rowstyle{\bfseries} Category & Prediction description & Example & Biomistral & Mixtral & LLaMa\\
		\hline
		& \\[\dimexpr-\normalbaselineskip+1pt]
		Format
		    & Followed the structure of a progress note
                & \makecell[l]{Assessment: \ldots
                                \\Plan: \ldots}
    	
    		& $83.4~(25)$ & $36.7~(11)$ & $96.7~(29)$ \\
        & \\[\dimexpr-\normalbaselineskip+1pt]
		\hline
        & \\[\dimexpr-\normalbaselineskip+1pt]
		Problem coverage
            & Captured the relevant problems or conditions of the patient
            & \makecell[l]{Respiratory Support: Given\ldots 
                            \\Anemia Management: The patient\ldots
                            \\$\vphantom{\int\limits^x}\smash{\vdots}$}
            & $76.7~(23)$ & $86.7~(26)$ & $96.7~(29)$ \\
        & \\[\dimexpr-\normalbaselineskip+1pt]
		\hline
        & \\[\dimexpr-\normalbaselineskip+1pt]
		\multirow{3.5}{*}{Copying}
            & Copied verbatim from the prior note
            & \makecell[l]{SEPTIC SHOCK- BPs low\ldots
                            \\(echoing prior note)}
            & $10.0~(3)$ & $0.0~(0)$ & $0.0~(0)$ \\
            & \\[\dimexpr-\normalbaselineskip+1pt]
		\cline{2-6}
            & \\[\dimexpr-\normalbaselineskip+1pt]
		& Copied verbatim from the summary of structured data
            & \makecell[l]{Their heart rhythm\ldots
                            \\(echoing summary)}
            & $13.3~(4)$ & $0.0~(0)$ & $0.0~(0)$ \\
            & \\[\dimexpr-\normalbaselineskip+1pt]
		\hline
            & \\[\dimexpr-\normalbaselineskip+1pt]
		\multirow{6}{*}{\makecell[l]{Relevant\\information}}
            & Updated the note with relevant information
            & {\def\arraystretch{1} \setlength\tabcolsep{1.5pt}
                \begin{tabular}{@{}p{0.42\linewidth}@{}}
    		        \textbf{Gold:} LFTs rising and will discontinue argatroban ? leperudin \\
    		        \textbf{Pred:} \ldots prothrombin time remains elevated at 31.5 seconds, and his INR is 3.2 \ldots\ anticoagulant regimen should be reviewed \ldots \\
    		    \end{tabular}
                }
            & $40.0~(12)$ & $73.3~(22)$ & $53.3~(16)$ \\
            & \\[\dimexpr-\normalbaselineskip+1pt]
		\cline{2-6}
            & \\[\dimexpr-\normalbaselineskip+1pt]
            & Updated the note leveraging the relevant structured data appropriately
              & {\def\arraystretch{1} \setlength\tabcolsep{1.5pt}
                \begin{tabular}{@{}p{0.42\linewidth}@{}}
    		        \textbf{Pred Summary:} prothrombin time elevated at 31.5 seconds, which indicates an increased risk of bleeding. His INR is also high at 3.2 \ldots\ may require adjustments to his anticoagulant regimen \\
    		    \end{tabular}
                }
            & \makecell{$76.9~(10)$\\out of $13$} & \makecell{$87.5~(21)$\\out of $24$} & \makecell{$100.0~(13)$\\out of $13$}  \\
        & \\[\dimexpr-\normalbaselineskip+1pt]
		\hline
        & \\[\dimexpr-\normalbaselineskip+1pt]
		Unrelated content
            & Included content that is unrelated to patient
            & \makecell[l]{\ldots hemoglobin increased to 7.5 g/dL\ldots
                            \\(absent in prior note or summary)}
            & $30.0~(9)$ & $23.3~(7)$ & $33.3~(10)$ \\
        	& \\[\dimexpr-\normalbaselineskip+1pt]
		\hline
	\end{tabular}
	}
\end{table}

\begin{figure}[t]
    \begin{center}
        \includegraphics[width=0.9\columnwidth]{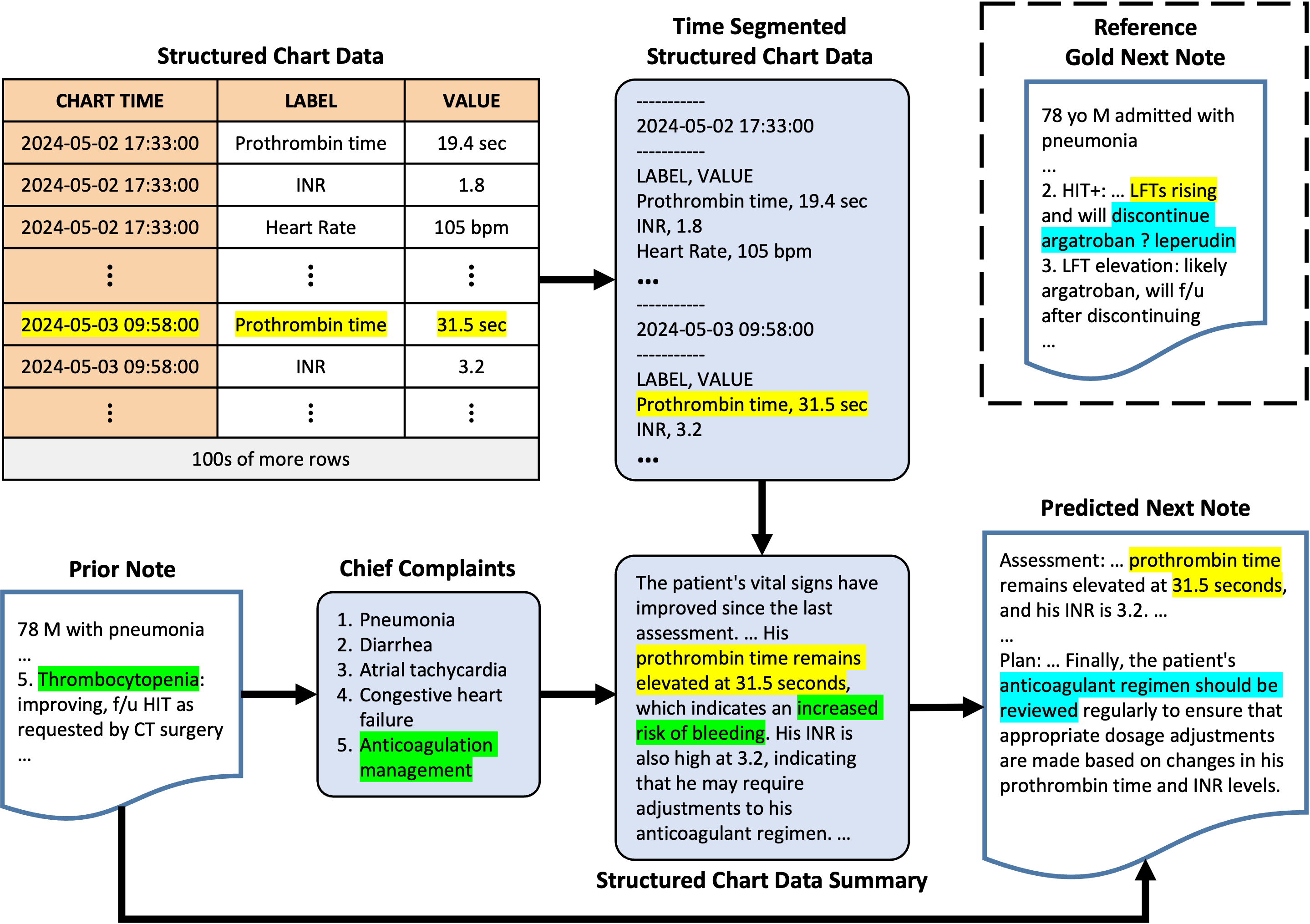} 
        \caption{A detailed example showing the flow of information in the proposed framework as the model makes intermediate and final predictions. The sample run is from the Biomistral model. For brevity and privacy restrictions related to the data source, some parts of the structured data, notes, and the predictions are omitted. Highlighted instances from \hlc[highlight-yellow]{assessment}, \hlc[highlight-green]{condition}, and \hlc[highlight-teal]{plan} illustrate information flow..}
        \label{fig:example-full}
    \end{center}
\end{figure}

The results from our manual evaluation are reported in Table \ref{tab:manual_eval_with_eg}.
The Biomistral model was apt at following the format of the A\&P note, outputting the note in an appropriate format $83.4\%~(25)$ of the time.
Interestingly, Mixtral was not able to capture the format as well as the other two models.
The list of patient's ongoing problems is captured with $76.7\%~(23)$ accuracy by Biomistral while the other larger models performed better.
All the models refrained from copying large chunks of text from the prior note and/or the structured data summary, with only the Biomistral model copying heavily in about $10\mhyphen 13\% (3\mhyphen 4)$ of cases.
Relevant information was identified and added to prediction in $73.3\%~(22)$ instances by Mixtral, which was the best among the models.
Interestingly, for Biomistral, $43.3\%~(13)$ of the generated summaries of structured data contained some relevant information to aid in generation.
In $10~(76.9\%)$ of these instances, the model took the available information into account.
Mixtral was able to extract important structured information the best (in $24$ instances) while LLaMa was the best in harnessing the extracted information ($100\%$ of the times).
We also encountered that all the models generated content which was not related to the patient in some instances, which can be considered as model hallucinations or confabulations.
A detailed example prediction from the Biomistral model is shown in Figure \ref{fig:example-full}.

\section{Discussion}

We described a task of automatically generating progress notes using structured patient chart data.
We present a novel framework and release a representative dataset, \dataset, for the task.
Our automated and manual analyses uncovers challenges associated with PNG, setting a jump-board for future research on generating progress notes.

The quality of generated summaries of structured EHR data played an important role in the proposed framework.
During our manual evaluation we saw that the summaries did not capture the relevant details in almost half of the cases.
Though we supplied the patient's chief complaints (e.g., \textit{``diabetes''}) to guide the summary generation, the model still sometimes picked up general clinical attributes (e.g., \textit{``vital signs''}) but ignored problem-specific attributes (e.g., \textit{``Glucose (serum)''}).
Clinical summarization is a well-studied task and there is potential to employ more advanced summarization techniques to improve PNG \cite{pivovarov2015AutomatedMethodsSummarization,keszthelyi2023PatientInformationSummarization}, especially given the modest context length of current medical LLMs.

There were also instances where the model could not leverage the available summary details (e.g., \textit{``oriented to own ability''}) to produce high-level assessment and plan components (e.g., \textit{``altered mental status has improved today''}).
Positively, the model included phrases such as \textit{``changes noted in respiratory rate''} for specific clinical attributes present in the summary but failed to highlight the nature of the change (e.g., \textit{``RR is high''}).
Another noteworthy observation is that the predicted notes usually contained broad drug classes (e.g., \textit{``anticoagulants''}) as opposed to specific medication names (e.g., \textit{``argatroban''}, \textit{``leperudin''}).
To this end, medical knowledge ingestion into the LLMs is a promising area for further exploration \cite{chen2023KnowledgeGraphsLife}.

Another interesting characteristic of the models was verbosity.
For almost all the analyzed instances, the model provided responses in proper English sentences that are atypical of the commonly seen telegraphic text in progress notes \cite{friedman2002TwoBiomedicalSublanguages}.
This behaviour seems agnostic to the semantics of generated content.
Regardless, the generated notes with familiar language may be more readily acceptable by the clinicians.
Thus, addressing these differences in writing style is an interesting area for further exploration.
The potential approaches for this include fine-tuning the models to the specific task with specialized data, which is known to adapt better to the target domain \cite{thirunavukarasu2023LargeLanguageModels}.

The automated evaluation metrics for medical note generation exhibit different trends on different data types \cite{benabacha2023InvestigationEvaluationMethods}.
In this case, the evaluation should ideally capture whether the model incorporates all relevant interim changes from the prior progress note while generating the next note.
We included this aspect in our manual evaluation, however, automating such judgements will greatly reduce the manual efforts required for assessing the quality of generated notes.

We view the proposed framework as an assistive tool for writing the progress notes, as opposed to a completely automated solution.
We use locally run models to alleviate data privacy concerns.
Further, bias mitigation is a critical aspect in medicine.
All applications of the proposed framework should take this into consideration and use the state-of-the art approaches to mitigate it.
Finally, for real-world applications, clinician in the loop should be educated to critically appraise algorithmic solutions.

We used structured EHR data and the prior progress note for PNG.
Since it is admissible for the model to have access to all prior information while generating the next progress note, other types of EHR data (e.g., radiology reports) can be explored as model input.
Due to the context length limitations, we employed an iterative framework to process the huge input.
Including more input data will likely require a larger model context size
or other modular approaches to get around such limitations.
Moreover, we used simple prompts to establish a baseline and found challenges associated with the proposed PNG task.
Incorporating more complex prompt engineering strategies (e.g., Chain-of-Thought \cite{wei2022ChainofthoughtPromptingElicits}, Retrieval-Augmented Generation \cite{lewis2020RetrievalaugmentedGenerationKnowledgeintensive}, Self-Consistency \cite{wang2023SelfconsistencyImprovesChain}) may help but was not the focus of this work.
Finally, LLMs fine-tuned on specific downstream natural language processing tasks have been shown to perform well \cite{wang2024DRGLLaMATuningLLaMA}.
To this end, the proposed dataset, \dataset, will be a useful resource to build models for generating clinical text.

\section{Conclusion}

We outlined the task of automated progress note generation using structured patient chart data.
The best performing LLM was Biomistral achieving a BERTScore F1 of $80.53$, underlining the importance of medical fine-tuning.
Through manual analysis, we surfaced challenges associated with the task and the shortcomings of the evaluated models.
We also raised concerns about the existing automated evaluation metrics and provided future directions for improving both the performance and evaluation for PNG.
Overall, we demonstrated that automating PNG is a feasible task, though further work is needed to address challenges identified during our analysis.

\section{Acknowledgments}

This work was supported by the intramural research program at the US National Library of Medicine, National Institutes of Health, and utilized the computational resources of the NIH HPC Biowulf cluster (\url{http://hpc.nih.gov}).

\bibliographystyle{vancouver-mod}
\setlength{\bibsep}{8pt}
\bibliography{png-amia24}

\end{document}